\documentclass[10pt,twocolumn,letterpaper]{article}

\usepackage{cvpr}              

\usepackage[dvipsnames]{xcolor}

\newcommand{\ts}[1]{\textsuperscript{#1}}

\usepackage{graphicx}
\usepackage{amsmath}
\usepackage{amssymb}
\usepackage{booktabs}
\usepackage{amsfonts}

\usepackage{float} 
\usepackage{array}
\usepackage{geometry}
\usepackage{fontawesome}
\usepackage{adjustbox}
\usepackage{tabularx}

\newcommand{\quoteArg}[1]{``#1''}

\newcommand{\Frst}[1]{\textcolor{red}{\textbf{#1}}}
\newcommand{\Scnd}[1]{\textcolor{blue}{\textbf{#1}}}

\newcolumntype{P}[1]{>{\raggedright\arraybackslash}p{#1}}

\definecolor{cvprblue}{rgb}{0.21,0.49,0.74}
\usepackage[pagebackref,breaklinks,colorlinks,citecolor=cvprblue]{hyperref}

\def\paperID{5} 
\def\confName{CVPR}
\def\confYear{2024}

\title{ARCH2S: Dataset, Benchmark and Challenges for Learning Exterior Architectural Structures from Point Clouds}

\author{Ka Lung Cheung\ts{1,2} \quad Chi~Chung~Lee\ts{2}\\ 
\ts{1}The Chinese University of Hong Kong \quad \ts{2}Hong Kong Metropolitan University\\
{\tt \small klcheung@mae.cuhk.edu.hk, cclee@hkmu.edu.hk}\\
\small \url{https://github.com/Semanticity-Research/ARCH2S}}

\begin{document}
\maketitle
\begin{abstract}
{Precise segmentation of architectural structures provides detailed information about various building components, enhancing our understanding and interaction with our built environment. Nevertheless, existing outdoor 3D point cloud datasets have limited and detailed annotations on architectural exteriors due to privacy concerns and the expensive costs of data acquisition and annotation. To overcome this shortfall, this paper introduces a semantically-enriched, photo-realistic 3D architectural models dataset and benchmark for semantic segmentation. It features 4 different building purposes of real-world buildings as well as an open architectural landscape in Hong Kong. Each point cloud is annotated into one of 14 semantic classes.}
\end{abstract}


\section{Introduction}


\begin{figure}
    \centering
    \includegraphics[width=1\linewidth]{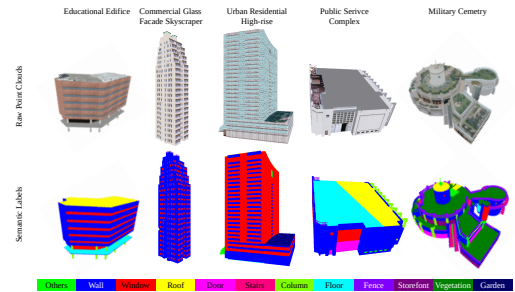}
    \caption{Point cloud models from ARCH2S dataset. Different colors label different semantic classes.  }
    \label{Fig:ARCH2S Dataset}
\end{figure}

Detailed segmentation of exterior architectural structures, such as fa\c{c}ades, elevations, and ground-level built objects, is crucial in various fields. It aids urban planning \cite{muhmad_kamarulzaman_uav_2023,gao_pssnet_2023}, improves efficiency in construction \cite{schonfelder_automating_2023,volk_building_2014,wiley_bim_2024,croce_semantic_2021,gu_understanding_2010}, informs restoration efforts in heritage preservation \cite{matrone_transfer_2021,pellis_image-based_2022,matrone_benchmark_2020}, and advances technologies such as autonomous vehicles \cite{biswas_autonomous_2023,bae_smart_2022,yao_radar-camera_2024} and drone navigation systems \cite{li_universal_2018,bolognini_vision-based_2022,alsayed_drone-assisted_2021,liang_towards_2023} by providing a semantic view of the surroundings \cite{selvaraju_buildingnet_2021}.


Despite advancements in LiDAR scanning technology and the rise of large-scale 3D point cloud datasets, capturing a comprehensive view of exterior architectural structures remains a challenge \cite{huang_city3d_2022,wang_building3d_2023}. Specifically, the outdoor roadway-level datasets \cite{yue_lidar_2018,hanselmann_king_2022,yao_radar-camera_2024,jiang_rellis-3d_2022,hu_towards_2021,loiseau_online_2022} often abstract the (large-sized) buildings as single entities, finer details of exteriors due to sensory gap \cite{gao_are_2020}. Although a recent surge of fa\c{c}ade-level datasets \cite{langlois_vasad_2022,whu_whu-tls_benchmark_dataset-1_nodate,wysocki_tum-faccade_2022} have aimed to address this problem by identifying the architectural parts on the building's vertical and planar wall surfaces, ground-level and rooftop-level exterior structures are disregarded. Considering the above data hunger problem \cite{gao_are_2020} and increasing adoption of Building Information Modeling (BIM)
\cite{schonfelder_automating_2023,tang_review_2019,darko_building_2020,kamel_review_2019}, a 3D architectural models dataset with rich annotations on every (level, angle) of the exterior surface is pressingly needed, which provides detailed categorization and distinction of different structural features.

In this paper, we present a richly annotated, semantically-enriched 3D \textbf{Arch}itectural models dataset -- ARCH2S. Our dataset (\cf \cref{Fig:ARCH2S Dataset}) includes \textbf{S}emantic \textbf{S}egmentation benchmark for learning exterior architectural structures. It contains real-world reconstruction scenes with photorealistic texture through texture mapping techniques. The dataset includes prominent buildings in Hong Kong and a historical open landscape. The significance of our dataset is summarized as follows:
\begin{itemize}
    \item Unlike the extensively researched outdoor roadway and fa\c{c}ade-level datasets, our dataset covers detailed annotations of real-world reconstructed exterior building objects into 13 commonly seen built elements grouped into three main categories (miscellaneous, structural, and decorative) (\cf \cref{fig:dataset_label_stat}). The broad spectrum of semantic classes creates fine-grained annotation on the fa\c{c}ades, roof, and ground-level objects.
    
    \item Being reconstructed from oblique aerial images, our point cloud models feature a polyhedral footprint and rooftop structures enhanced with photorealistic textures. Meanwhile, our dataset is evaluated with the state-of-the-art algorithms in the 3D segmentation task. Through evaluation of the annotation (\cf \cref{fig: Labeling errors}) and benchmark results (\cf \cref{tab:benchmark_comparison}), we discuss the challenges (\cf \cref{pgh: Learning Exterior Architectural Structures}) in annotation and semantic segmentation for large-sized 3D architectures.
\end{itemize}

\section{The ARCH2S Dataset}
\begin{figure}[t]
    \centering
    \includegraphics[width=0.92\linewidth]{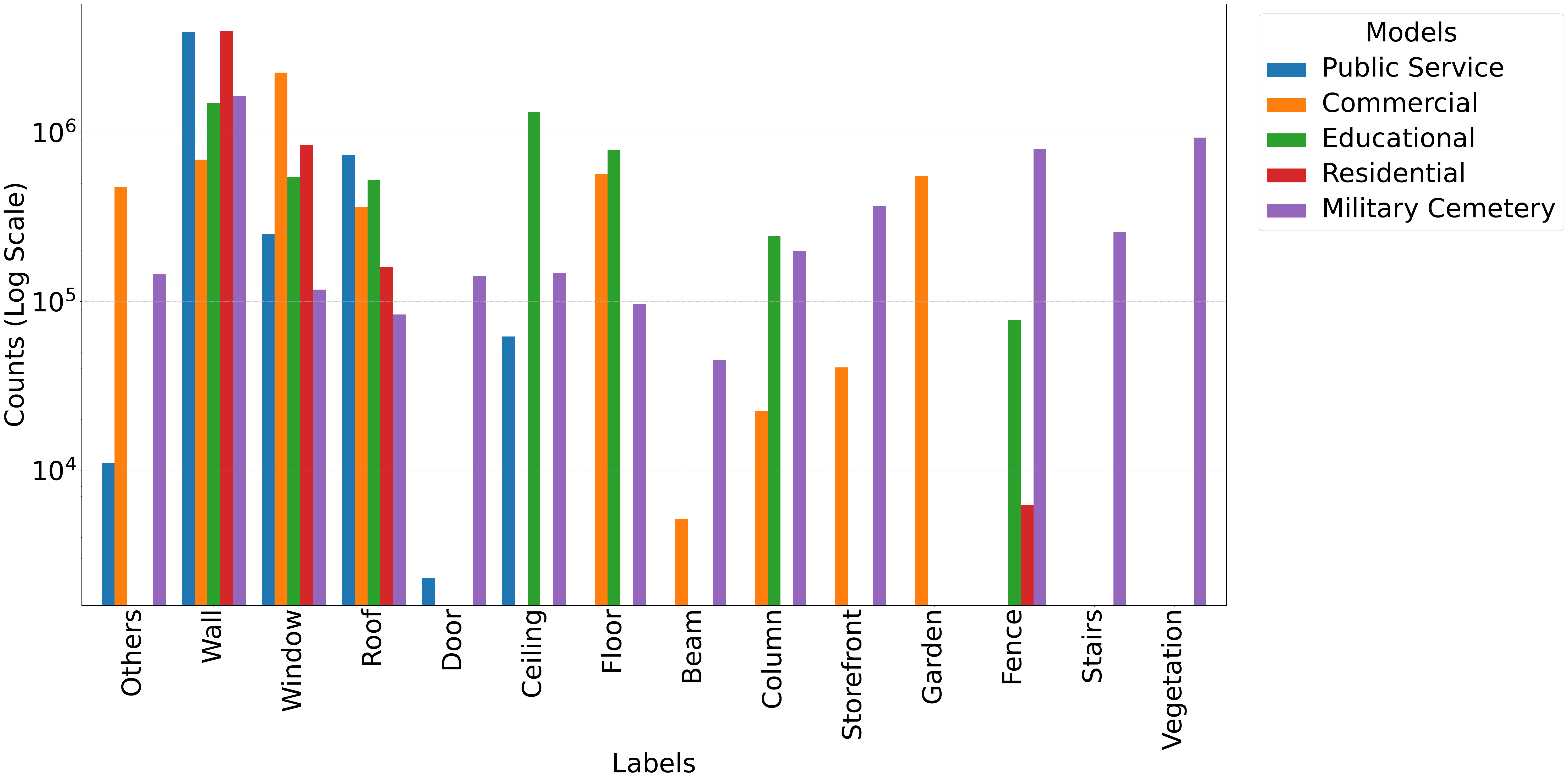}
    
    \caption{The distribution of different semantic categories in ARCH2S dataset. Our dataset, consisting of 14 out of 22 semantic classes, covers a diverse range of common exterior architectural elements. Note that the \quoteArg{Garden} semantic label is reported only in the Commercial Centre point cloud. Moreover, point counts are on a logarithmic scale for the vertical axis.  }
    
    \label{fig:dataset_label_stat}
\end{figure}

Our dataset contains prominent buildings and an open landscape from Hong Kong for five different architectural purposes: educational, residential, commercial, public service, and military cemetery (\ie the open landscape) architecture. We randomly sampled approximately 5M points for each scene from a mesh object source to generate point cloud models. These models are annotated with 13 commonly seen structural and decorative built object classes, along with an extra class (\quoteArg{Others}) for architectural components that do not fit into the predetermined 13 classes (\cf \cref{fig:dataset_label_stat}).
\subsection{Dataset Preparation}
\begin{figure}[t]
    \centering
    \includegraphics[width=1\linewidth]{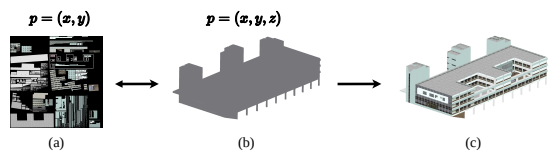}
    \caption{Illustration of the stages of UV mapping with our mesh (Educational Facility). \textbf{(a)} The UV layout with 2D coordinates $(x, y)$, representing each point $p$ on the mesh's surface. \textbf{(b)} The 3D mesh in gray-scale, showing points $p$ in 3D space ($x, y, z)$ for texture projection. \textbf{(c)} The final textured mesh, with the UV map applied, displays detailed and colored surfaces.}
    \label{fig: UV-mapping}
\end{figure}

\begin{figure}[t]
    \centering
    \includegraphics[width=0.7\linewidth]{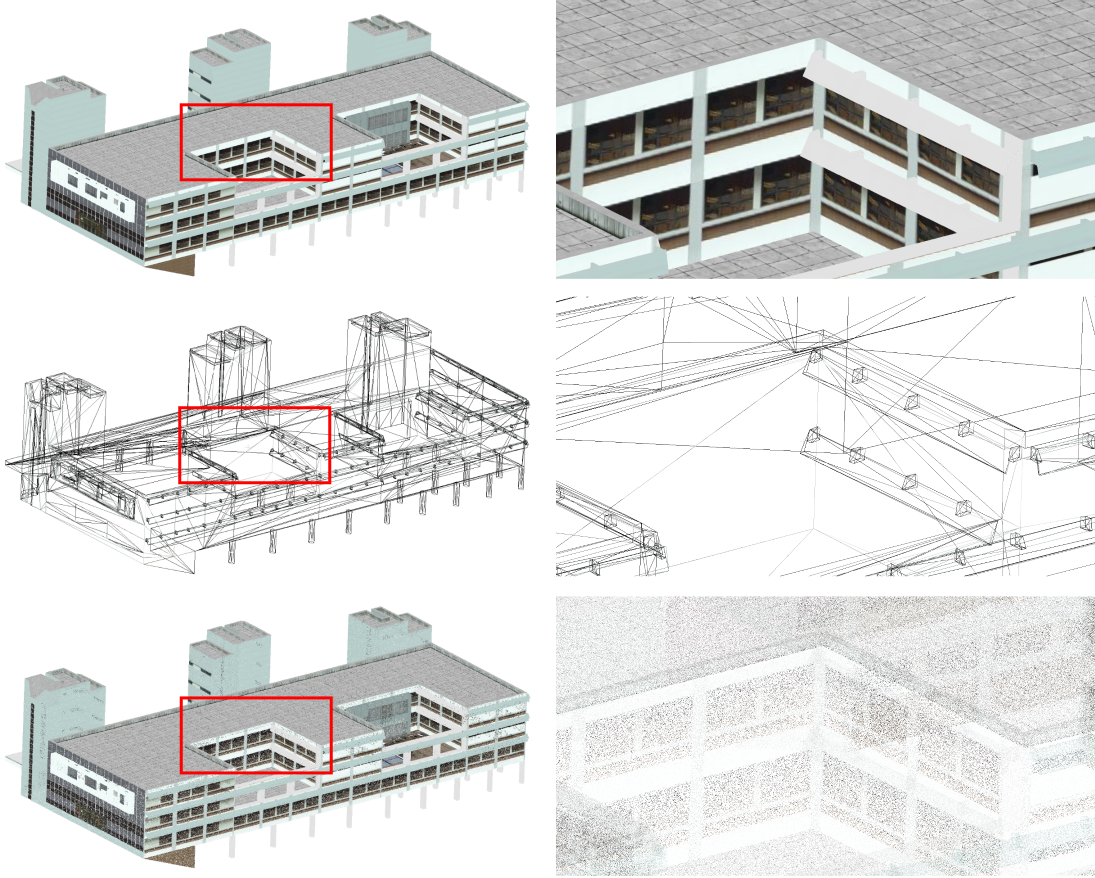}
    \caption{Illustration of the random point sampling on our mesh. Plain mesh (top), wireframe display (middle), and corresponding sampled points (bottom).}
    \label{fig: Educatinal-facility-zoom-in}
\end{figure}

\begin{figure}[t]
    \centering
    \includegraphics[width=1\linewidth]{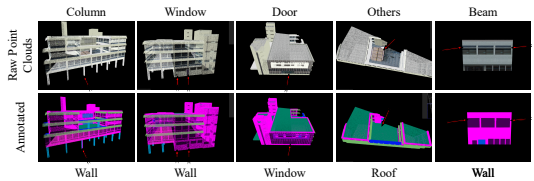}
    \caption{Comparative analysis of raw point cloud data and semantic annotations in the Educational Facility, visualized on the annotation interface. The top row displays raw point cloud representations, while the bottom row shows the semantically annotated elements with various colors denoting different labeled classes. \Frst{Red} arrows indicate discrepancies where the semantic labels do not accurately match the actual built elements, highlighting areas of potential mislabeling.}
    \label{fig: Labeling errors} 
\end{figure}

\paragraph{Models Mining.}
Our 3D models are derived from the 3DBIT00 dataset \cite{noauthor_3d_2024} provided by the Hong Kong Lands Department \cite{lands_department_-_the_government_of_the_hong_kong_special_administrative_region_lands_2024}. The dataset is updated bi-monthly and features individualized 3D models with geometric shapes, appearance, and position of three types of ground objects: buildings, infrastructure, and terrain. Our project extracted targeted FBX formatted models (mesh models), which were to be annotated by human operators. Additionally, we selected an extra Educational Facility model (\cf \cref{fig: UV-mapping,fig: Labeling errors,fig: Educatinal-facility-zoom-in}). This served as a sample for annotator training and to benchmark annotators' performance. 


\paragraph{Mesh Pre-processing.}
The FBX building meshes were provided with a UV texture map. This map assigns a 2D image to a set of coordinates, where U represents the horizontal axis, and V represents the vertical axis. By employing UV mapping(\ie texture mapping \cite{catmull_e_subdivision_1974,tecgraf_institute_of_puc-rio_5_2019}),, as shown in \cref{fig: UV-mapping} (a-c), the object appearances (\eg color, texture, and pattern) were incorporated into the surface of our 3D meshes.




\paragraph{Point Sampling from Mesh.}
After completing the UV mapping, we generated a point cloud on the UV-mapped 3D mesh, shown in \cref{fig: Educatinal-facility-zoom-in}. This was achieved by randomly sampling (5M) points on the surface of the mesh model. We set a target of 5M sampling points to maintain data granularity while minimizing data volume to ensure a successful loading into the annotation interface. The color information from the original mesh was collected by interpolating this data within each triangle. Each randomly positioned point adopted the properties of the UV coordinates at its location, thus preserving the visual details from the texture map.


\begin{table*}[t]
\centering
\caption{Quantitative results of five selected baselines on ARCH2S dataset Overall Accuracy (OA, \%), mean class Accuracy (mAcc, \%), mean IoU (mIoU, \%), and per-class IoU (\%) are reported. Miscellaneous: built element that belongs inclusively into decorative and structural categories. All baselines are trained from scratch (without pre-trained weights). The best and second-best results within each metric are denoted in \Frst{red} and \Scnd{blue}, respectively.}
\label{tab:benchmark_comparison}
\resizebox{\textwidth}{!}{
\begin{tabular}{@{}l*{3}{c}*{6}{c}*{5}{c}*{3}{c}@{}}
\toprule
 & & & &  \multicolumn{7}{c}{\textbf{Miscellaneous}} & \multicolumn{4}{c}{\textbf{Structural}}   & \multicolumn{3}{c}{\textbf{Decorative}}\\ 
\cmidrule(lr){5-11} \cmidrule(lr){12-15} \cmidrule(lr){16-18}
 Methods & \rotatebox{90}{OA (\%)} & \rotatebox{90}{mAcc (\%)} & \rotatebox{90}{mIoU (\%)} & \rotatebox{90}{Others} & \rotatebox{90}{Wall} &  \rotatebox{90}{Window} & \rotatebox{90}{Door} & \rotatebox{90}{Roof} &  \rotatebox{90}{Storefront} & \rotatebox{90}{Ceiling}  & \rotatebox{90}{Floor} & \rotatebox{90}{Beam} & \rotatebox{90}{Stairs} & \rotatebox{90}{Column}  & \rotatebox{90}{Fence} & \rotatebox{90}{Vegetation} & \rotatebox{90}{Garden}  \\
\midrule
SpUNet & 30.98 & \Scnd{14.8} & \Scnd{6.45} & \Scnd{0.79}  & \Frst{21.75} & \Scnd{24.66} & 0.00 &  \Scnd{24.77} & \Frst{32.18}  & 0.00  & 0.00 & 0.00  & 0.00 & \Frst{18.03} & 0.00 & 0.00 & 0.00\\
MinkUNet & \Scnd{32.31} & \Frst{15.28}  & \Frst{7.39} & 0.00 & \Scnd{20.43} & 7.5 & 0.00  & \Frst{38.98} & 13.45 & 0.00 & \Frst{60.01} & 0.00 & 0.00 & 0.00  & 0.00 & 0.00 & 0.00 \\
\midrule
 PTv1  & \Frst{48.64} & 5.94 & 3.44 & \Frst{7.42} & 6.06 & \Frst{51.91} & 0.00 & 0.00 & 0.00 & 0.00 & 0.00 & 0.00 & 0.00 & 0.00 & 0.00 & 0.00 & 0.00 \\
 PTv2 & 18.50 & 7.95 & 3.16 & 0.00 & 18.72 & 0.94 & 0.00 & 0.00  & \Scnd{16.84}  & 0.00 & \Scnd{23.57}  & 0.00 & 0.00 & 0.00 & 0.00 & 0.00 & 0.00  \\
PTv3 & 22.59 & 10.07 & 3.01 & 0.09  & 19.91 & 8.05 & 0.00 & 24.30 & 4.85 & 0.00  & 0.00 & 0.00  & 0.00 & 0.00 & 0.00 & 0.00 & 0.00\\
\bottomrule
\end{tabular}
}

\end{table*}


\subsection{Benchmarks}  

\paragraph{Train and Test Splits.}
We define standard training and testing splits so that no duplicates occur between train and test splits. We first trained with Educational, Residential, and Public Service buildings and Military cemetery and tested on Commercial buildings. Each building model is downsampled into grid boxes of $0.75$ x $0.75$ m and hashed with voxel hashing \cite{xu_multi-scale_2018}, which includes the global geometries and RGB values for point clouds.

\paragraph{Representative Baselines.}
Neural segmentation methods in scene semantic segmentation include, but are not limited to, two mechanisms: convolution-based  \cite{spconv_contributors_spconv_2022,choy_4d_2019} and the recent point-wise transformer-based \cite{zhao_point_2021,wu_point_2022,wu_point_2023} methods. We have selected five representative methods as solid baselines for benchmarking our dataset. The candidate models from convolution-based methods include SpUNet \cite{spconv_contributors_spconv_2022,choy_4d_2019} and MinkUNet \cite{choy_4d_2019}, while transformer-based methods include  PTv1 \cite{zhao_point_2021}, PTv2 \cite{wu_point_2022}, and PTv3 \cite{wu_point_2023}.

\paragraph{Evaluation Metrics.} Similar to existing benchmarks \cite{hackel_semantic3dnet_2017,behley_semantickitti_2019}, we use Overall Accuracy (OA) and mean Intersection-over-Union (mIoU) as the primary evaluation metrics.

\section{Results and Discussion} \label{Sec: Results and Discussion}

\paragraph{Challenges in Annotation.}
Despite providing the training and instructions to the annotators, We report a few potential mislabeled objects in the annotated model generated by the human operators (\cf \cref{fig: Labeling errors}) compared to the raw point cloud scene in the training phase. The inconsistencies in labeling are intelligible, considering the semantic gaps from annotators \cite{gao_are_2020}. Annotators have different viewpoints or ambiguous definitions when classifying the building components (\eg classes: \{\quoteArg{Column}, \quoteArg{Beam}, \quoteArg{Wall}\}, classes: \{\quoteArg{Floor}, \quoteArg{Fence}, \quoteArg{Window}\}, classes: \{\quoteArg{Roof}, \quoteArg{Ceiling}\}). 

\paragraph{Learning Exterior Architectural Structures.} \label{pgh: Learning Exterior Architectural Structures}
\cref{tab:benchmark_comparison} shows the convolutional methods outperform Vision Transformers (ViTs) in terms of mIoU across various categories. SpUNet \cite{spconv_contributors_spconv_2022,choy_4d_2019} and MinkUNet \cite{choy_4d_2019} are able to learn richer local features on neighborhood points and handle the sparsity of large-scale point clouds efficiently \cite{guo_deep_2020,he_deep_2023}, resulting with greater generalization compared to PTv1 \cite{zhao_point_2021}, PTv2 \cite{wu_point_2022}, and PTv3 \cite{wu_point_2023}. Conversely, ViTs result poorly primarily due to the lack of locality, inductive biases, and hierarchical structure of the representations \cite{tay_efficient_2022,liu_efficient_2021,lu_transformers_2022,guo_deep_2020} when processing with large data grids. Thus, high-capacity models like ViTs are seen to be less efficient when trained with our dataset.

\paragraph{Impact of Imbalanced Semantic Distribution.} \label{pgh:impact of imbalanced semantic distribution}
The segmentation performance in building classes varies, with dominant classes performing better than underrepresented, minor classes. Major structures like \quoteArg{Wall} and \quoteArg{Window} result in a higher per-class IoU score (\cf \cref{tab:benchmark_comparison}) than the minority, suggested by their scene occupation and class dominance. These class objects appear to have large planar surfaces (\ie large amount of points on the same class), which bring scant novel information to scene understanding. In contrast, segmenting those minority classes (\eg \quoteArg{Beam} and \quoteArg{Column}) is challenging due to under-representation, leading to less accurate segmentation (\ie significantly low or nearly zero per-class IoU). Additionally, minor class objects inherently have more complex, varied shapes, styles, and patterns than other major class objects, which poses additional challenges to model generalization. To improve the semantic learning of the building model, transfer learning \cite{abu_dabous_condition_2020} or different class-weighted loss functions \cite{abu_dabous_condition_2020,hu_towards_2021} are possible solutions to mitigate the impact of long-tail data distribution \cite{gao_are_2020}.

\section{Conclusion}
In this paper, we introduced a semantically-enriched 3D architectural models dataset with varied semantic classes for the commonly seen building components. Through our analyses of annotations and benchmarking results, We identified several open challenges: 1) inconsistent annotations for complex architecture due to the semantic gap, 2) the issue of model generalization, and 3) the impact of the imbalanced class distribution. Despite these challenges, our dataset aspires to be pioneering work to encourage the adoption of BIM and novel applications related to smart cities. 


\newpage

{\small
\bibliographystyle{ieeenat_fullname}
\bibliography{references4}
}

\end{document}